\documentclass[11pt]{article}

\usepackage[margin=1.1in]{geometry}
\usepackage{amsmath}
\usepackage{amssymb}
\usepackage{booktabs}
\usepackage{enumitem}
\usepackage[hidelinks]{hyperref}
\usepackage{lmodern}
\usepackage[T1]{fontenc}
\usepackage{microtype}
\usepackage{xspace}
\usepackage{tikz}
\usetikzlibrary{arrows.meta,positioning,fit}
\tikzset{
  stage/.style={draw, rounded corners=1pt, align=center,
    font=\small, inner sep=4pt, minimum height=1.9em},
  note/.style={font=\footnotesize\itshape, align=center},
  flow/.style={-{Stealth[length=2mm]}, semithick},
  refused/.style={-{Stealth[length=2mm]}, semithick, dashed},
}

\newcommand{\sys}{Quipu\xspace}
\newcommand{\census}{Census\xspace}
\newcommand{\entails}{\models}
\newcommand{\Sig}{\ensuremath{\Sigma}\xspace}

\title{\textbf{\sys: A Governed Bitemporal Knowledge Graph Store}\\
\large Start strict: rethinking knowledge-graph defaults for agent-written knowledge}
\author{Steve Brown\thanks{\textsc{orcid} \href{https://orcid.org/0009-0009-1720-1785}{\nolinkurl{0009-0009-1720-1785}}. With implementation and drafting assistance
from Claude (Anthropic). \textbf{Artifact availability:} the source and
the \texttt{benchmark/census/} artifacts behind every number reported
here are archived at \textsc{doi}
\href{https://doi.org/10.5281/zenodo.21878429}{\nolinkurl{10.5281/zenodo.21878429}}
(v0.3.20)~\cite{brown2026quipu}; the concept \textsc{doi}
\href{https://doi.org/10.5281/zenodo.21878428}{\nolinkurl{10.5281/zenodo.21878428}}
resolves to the newest release. Development repository:
\texttt{github.com/scbrown/quipu}.}}
\date{\today}

\begin{document}
\maketitle

\begin{abstract}
Agents now write knowledge graphs, but knowledge-graph stores still
carry defaults set when humans curated them: accept writes now and
clean later, keep one time axis or none, treat every writer's facts as
equally trustworthy, and leave governance to dashboards and middleware.
We argue these four defaults are individually convenient and jointly
untenable under agent workloads, and we present \sys, an embeddable
store that inverts all four: no fact enters except through a gate whose
predicates evaluate the pending \emph{post-state}; data, trust labels,
verdicts, and the rules themselves are bitemporal; named graphs are the
unit of authority and trust, composed under a lattice whose single
invariant is that composition never widens; and the governance
specification \Sig, the trace, and signed verdicts are facts in the
store they govern, making the audit $T \entails \Sig$ a query.
We evaluate with \census, a deterministic multi-writer lifecycle whose
single seeded run scores every research question against planted ground
truth: the gated store ends with 0 of 6 planted defects versus 6 of 6
ungated; all 7 composition probes uphold the lattice contract;
50 of 50 satisfied verdicts re-derive faithfully as of their instant
while all 50 would be misreported under a latest-only rule set; and the
SARC reference checker agrees with the in-store audit verdict-for-verdict,
disagreeing only on coverage semantics. A recorded trace from a real
governed writer surfaces a live enforcement gap the audit names with
its remediation. Against an external decision-evidence sufficiency
benchmark (DEMM-Bench), a content-only reading of the store's exported
records answers all 512 property-level governance questions correctly
with zero overclaim under all eight degradation conditions, while
container-presence baselines overclaim on up to 87.5\% of the same
cases --- and the run itself surfaced, and led us to close, a gap in
what a denial's verdict attests.
\end{abstract}

\section{Introduction}
\label{sec:intro}

Knowledge graphs are increasingly written by software agents: language
models extracting facts from documents, code-analysis engines promoting
structural facts, pipelines reconciling external corpora. The stores
underneath them, however, retain defaults chosen for a different
author. When a human curator is the writer, it is reasonable for a
store to accept whatever arrives and rely on later review; to keep at
most one time axis; to treat all writers alike; and to leave policy to
the perimeter. When the writer is an agent that produces plausible,
well-formed, and sometimes wrong facts faster than any review process
drains them, each of those defaults becomes a liability.

We name four defaults and their failure modes
(Table~\ref{tab:defaults}). \textbf{D1 --- accept, then clean:}
validation in conventional stores is optional, post-hoc, or the
application's job; with agent writers the cleanup debt compounds and
downstream readers consume the store in the window between write and
review. \textbf{D2 --- one time axis, or none:} even honestly bitemporal
stores make only the \emph{data} bitemporal, so ``what was trusted at
time $T$'' and ``what was \emph{allowed} at $T$'' are unanswerable, and
audit degrades to log archaeology. \textbf{D3 --- flat trust:} named
graphs exist, but composition is silent union; one query joins an
attested graph with a quarantined one and the result inherits the
prestige of the attested source. \textbf{D4 --- governance outside:}
policy lives in dashboards, prompts, or middleware; the specification
and the store drift independently, and no mechanical check connects
them.

\sys is an embeddable knowledge-graph store that inverts all four
defaults, and this paper's claim is the \emph{conjunction}: each
inversion exists somewhere in the literature, but no store ships all
four, and the four reinforce each other. The gate is worth trusting
because its verdicts are permanent, signed, bitemporal facts; the
label lattice is enforceable because partitions gate authority; the
audit is decidable because governance is data; and bitemporality makes
all of it historical rather than merely current.

\begin{table}[t]
\centering
\small
\begin{tabular}{@{}llll@{}}
\toprule
 & Conventional default & \sys's inversion & Measured by \\
\midrule
D1 & accept, then clean & refuse at the gate; agents retry & RQ1, RQ2 \\
D2 & one time axis, or none & data, labels, verdicts, rules bitemporal & RQ5 \\
D3 & flat trust & partitioned trust; non-widening composition & RQ4 \\
D4 & governance outside & \Sig, trace, verdicts as facts; audit is a query & RQ3 \\
\bottomrule
\end{tabular}
\caption{The four defaults and their inversions.}
\label{tab:defaults}
\end{table}

The operating posture that falls out is the thesis in one line:
\emph{start strict; use agents to bear the cost of strictness.} A store
that refuses invalid, untagged, unauthorized, or policy-violating
writes is affordable exactly when the writer is an agent, because the
refusal carries structured feedback and the agent --- not a curator ---
absorbs the retry (\S\ref{sec:eval-agent} shows the loop converging in
one revision).

\paragraph{Contributions.}
\begin{enumerate}[leftmargin=1.4em,itemsep=0.15em]
\item \textbf{The system} (\S\ref{sec:datamodel}--\S\ref{sec:impl}):
  a single-file, embeddable store implementing all four inversions ---
  a bitemporal EAVT log with a three-valued write operation,
  named-graph partitioning with bind-once overlays and tombstones, a
  machine-checked label lattice with explicit coverage, and an in-store
  governance plane with signed verdicts, escalation, authority
  intersection, and a deterministic audit.
\item \textbf{The design principles} (\S\ref{sec:principles}): GS1--GS6,
  a one-page statement of what a store must guarantee before
  agent-written knowledge can be trusted, each principle paired with
  the failure mode of the conventional default it replaces --- distilled
  from building \sys, and extending SARC's governance-by-architecture
  from the agent loop to the store.
\item \textbf{The benchmark} (\S\ref{sec:census}): \census, a
  deterministic multi-writer lifecycle whose single seeded run measures
  all four inversions against planted ground truth, with byte-identical
  manifests across repeat runs, plus an in-the-wild replay of a real
  governed writer's trace.
\end{enumerate}

\section{Background and Requirements}
\label{sec:background}

\paragraph{Governance by architecture.}
SARC~\cite{besanson2026sarc} treats constraints as first-class
specification objects alongside state, action space, and reward: a
constraint declares its source, class (\emph{hard} / \emph{soft} /
\emph{escalation}), predicate, verification point, and response, and
compiles into four enforcement points in the agent loop --- a Pre-Action
Gate, an Action-Time Monitor, a Post-Action Auditor, and an Escalation
Router --- bound by invariants whose joint effect is a \emph{decidable
audit}: given a specification \Sig with constraint set $C$ and a
trace $T$, a checker decides $T \entails \Sig$ --- read: the trace
satisfies the specification --- in $O(|T|\cdot|C|)$, trace records
times constraints, without access to the model or its prompts. SARC stops at the loop: its reference artifact keeps \Sig
in a file beside the system and audits an exported trace with an
external checker. The knowledge the agents act on sits in an
ungoverned store underneath. Our position is that the store is the
right compilation target for exactly the machinery SARC specifies.

\paragraph{Bitemporal data.}
A bitemporal store indexes facts by transaction time (when the store
learned something) and valid time (when it holds in the world), in the
lineage of Datomic and XTDB. What conventional bitemporal systems do
not do is extend the treatment beyond data: trust annotations, policy
decisions, and the validation rules themselves remain latest-only, so
the historical questions governance actually asks --- what was believed,
what was trusted, what was \emph{required} at $T$ --- outrun the store's
memory.

\paragraph{Named graphs and trust.}
RDF datasets partition triples into named graphs, and the provenance
literature attached trust semantics to them two decades
ago~\cite{carroll2005named}; information-flow lattices are older
still~\cite{denning1976lattice}. SPARQL's dataset construction,
however, is silent union: nothing in the standard machinery refuses a
composition, degrades it, or even remarks that one member graph carries
no declaration at all.

\paragraph{Requirements.}
A store serving multi-writer agent ingestion in a governed setting
needs, at minimum: write-time constraint evaluation against the state
the write would create; a permanent, attributable record of every
gate decision, including refusals; authority that attaches to
partitions and only narrows under delegation; composition that cannot
launder trust; an audit decidable from the store's own contents; and
reproducibility of past decisions under the rules in force at the
time. \S\ref{sec:principles} states these as six principles;
\S\ref{sec:datamodel}--\S\ref{sec:governance} show the mechanisms that
discharge them.

\section{Design Principles: the Governed Store}
\label{sec:principles}

Six principles, distilled from building \sys, each paired with the
conventional failure it prevents. Nothing below names SQLite, RDF, or
EAVT --- the principles are substrate-agnostic on purpose, and stating
them as a portable contract for other substrates is future work
(\S\ref{sec:conclusion}).

\begin{description}[leftmargin=1.4em,itemsep=0.35em]
\item[GS1 --- Gated writes.] No fact enters the store except through a
  gate whose predicates evaluate against the pending
  \emph{post-state}. Post-state, not pre-state and not the request: a
  constraint like ``no entity holds two placements'' is checkable only
  after the candidate facts are staged --- a pre-state gate passes a
  write that is valid alone and invalid in combination
  (\census probe CEN-P2 separates the two). Writes touching no
  governed target must incur no policy-evaluation cost, or strictness
  becomes an argument against adoption. \emph{Prevents:} D1's
  accept-then-clean debt.
\item[GS2 --- Verdict permanence.] Every gate outcome --- allow, deny,
  and unknown --- persists as a signed, time-indexed fact that survives
  rollback of the write it judges. The ordering is the content: a
  denied write is rolled back, so verdicts are staged outside the
  write's transaction and flushed after it resolves --- the denial's
  verdict is precisely the record worth keeping. No signing identity
  means no verdict, never an unsigned one; signatures verify against a
  human-authored root of trust the store cannot mint for itself.
  \emph{Prevents:} unauditable refusal --- the conventional store throws
  away exactly the events an auditor needs most.
\item[GS3 --- Partitioned authority.] Authority attaches to partitions;
  delegation only narrows (intersection along the principal chain,
  with the wildcard as identity); an empty intersection refuses,
  never falls back. Changing a partition's standing requires authority
  over the \emph{meta}-partition --- otherwise a tenant promotes itself
  to attested. Overlays bind once to their parent, so a layer cannot
  forge presence in a base it was never bound to. \emph{Prevents:}
  D3's flat-trust escalation.
\item[GS4 --- Non-widening composition.] A view composed from
  partitions carries a label no stronger than the fold of its parts.
  Freshness and trust fold by meet; obligations by join (one
  \texttt{no-export} member taints the set). Undeclared is not a
  lattice value: the composed result is a pair of fold and
  \emph{coverage}, and partial coverage fails enforcement floors ---
  fail-safe at enforcement, honest at reporting. Trust from different
  declared chains refuses comparison by name rather than ordering
  silently. Expiry is absence, not falsity. \emph{Prevents:} trust
  laundering through views.
\item[GS5 --- In-store decidable audit.] \Sig, the trace $T$, and the
  verdicts live in the store they govern; $T \entails \Sig$ decides in
  $O(|T|\cdot|C|)$ without the model or its prompts, and a trace that
  \emph{contradicts} \Sig (violation) is never conflated with a trace
  that \emph{under-determines} it (incompleteness) --- the two demand
  different responses, and an audit that merges them invites both
  being ignored. \emph{Prevents:} D4's governance-by-log-grepping.
\item[GS6 --- As-of replay.] Every governance decision is reproducible
  against the store as of its transaction --- the facts, the labels,
  \emph{and the rules} in force at the time. The rules half is the
  hard half: a store whose shapes and policies are latest-only can
  replay what was known but not what was required, and a mid-lifecycle
  amendment makes the difference observable
  (\S\ref{sec:eval-replay}). \emph{Prevents:} D2's unanswerable
  ``what was allowed when.''
\end{description}

The principles interlock rather than stack: GS1's gate is worth
trusting because GS2 makes its outcomes permanent evidence; GS2's
verdicts are believable because signing verifies against a root of
trust GS3 keeps out of the store's own hands; GS4's lattice is
enforceable because GS3 makes partitions the unit a floor can refuse;
GS5's audit is decidable because GS1--GS2 already produced
\Sig-shaped traces and verdicts as data; and GS6 makes every other
guarantee historical rather than merely current. This interlocking is
why the contribution is the conjunction, not the parts.

\section{Data Model}
\label{sec:datamodel}

\paragraph{Bitemporal EAVT with a three-valued operation.}
The substrate is an append-only fact log
$(e, a, v, g, tx, \mathit{valid\_from}, \mathit{valid\_to}, op)$:
entity, attribute, value, named graph, transaction, valid interval,
and operation. Transaction time comes from a transactions table
carrying actor and source; current state is the open-interval subset.
IRIs are interned in a term dictionary whose append-only discipline is
itself an invariant under test. The operation is three-valued:
\emph{assert}, \emph{retract} (a logical close of the valid interval ---
history survives), and \emph{tombstone}, which marks a triple
\emph{absent in a composed view} without mutating the layer beneath ---
an operation neither Datomic-style logs nor RDF carry, and the piece
that makes layered composition sound.

\paragraph{Named graphs, overlays, datasets.}
The graph coordinate $g$ partitions the log. Committed graphs are
self-rooted; overlay-class graphs bind \emph{once} at creation to a
committed parent branch, and rebinding is an error --- the binding is
unforgeable, which is what lets a composed view resolve
nearest-overlay-wins without trusting the overlay's claims about its
base. Datasets name arbitrary graph-sets; the branch tree and the
dataset semilattice are different relations over the same nodes, and
silence never widens a dataset.

\paragraph{The label lattice.}
Partitions carry labels on four axes --- freshness, trust, durability,
policy --- stored as ordinary bitemporal facts in a reserved meta-graph
(the cache columns are derived, re-computable state). Composition
folds labels under one named invariant, \emph{composition never
widens}: freshness and trust by meet, obligations by join. The
composed result is a pair $(\mathit{fold}, \mathit{coverage})$ with
coverage in $\{\textsf{Empty}, \textsf{None}, \textsf{Partial},
\textsf{Full}\}$; \textsf{Empty} is the fold identity and distinct
from \textsf{None}. The homomorphism
$\mathit{label}(A \cup B) = \mathit{label}(A) \sqcap \mathit{label}(B)$
is machine-checked by property test. Trust ranks compare only within a
declared chain; cross-chain comparison returns an error naming both
chains, because a silent integer comparison is exactly the bug that
ranks a learned tactic above canon.\footnote{The lattice's motivating
deployment: NeuralAmplifier
(\texttt{github.com/scbrown/NeuralAmplifier}), a game-playing harness
whose knowledge divides into exactly such planes --- canonical rules
(datalinks), curated doctrine, learned memory --- and whose correctness depends
on the precedence between them never being decided by accident.} Label expiry is a
$\mathit{valid\_to}$ on the label assertion: an expired label is
absent --- it degrades coverage --- not false and not unknown.

\paragraph{Composition across stores.}
Term identifiers are globally unique by construction (each store
allocates from its own space, $s \cdot 2^{40} + k$), because reference
values are opaque payloads no query-time remapping can rewrite.
Read-only attachment mounts another store as a composed layer under
three invariants: attaching changes no existing query's result, the
host never writes into an attachment, and attachments are verified and
refused rather than migrated. Knowledge packs move a graph between
stores by \emph{re-interning} facts rather than copying rows, and a
pack's identity is a content hash over sorted N-Triples --- two stores
with different id assignment hash the same.

\section{The Governance Plane}
\label{sec:governance}

\begin{figure}[t]
\centering
\resizebox{\textwidth}{!}{%
\begin{tikzpicture}[node distance=4mm and 8mm]
  \node[stage] (write) {write\\\footnotesize datums, instant,\\\footnotesize actor, chain, graph};
  \node[stage, right=of write] (auth) {authority\\\footnotesize intersection\\\footnotesize along chain (GS3)};
  \node[stage, right=of auth] (gate) {policy gate\\\footnotesize \Sig claims (\textsc{ask})\\\footnotesize over staged post-state};
  \node[stage, right=of gate] (shacl) {SHACL\\\footnotesize episode\\\footnotesize shapes};
  \node[stage, right=of shacl] (commit) {commit\\\footnotesize bitemporal\\\footnotesize facts};
  \node[stage, below=7mm of gate] (rollback) {rollback\\\footnotesize attempted delta\\\footnotesize discarded (GS2)};
  \node[stage, below=16mm of commit, xshift=2mm] (verdict) {signed verdict\\\footnotesize policy, target, outcome,\\\footnotesize actor + chain in sealed hash};
  \draw[flow] (write) -- (auth);
  \draw[flow] (auth) -- (gate);
  \draw[flow] (gate) -- (shacl);
  \draw[flow] (shacl) -- (commit);
  \draw[refused] (auth) |- node[note, pos=0.75, above] {refuse} (rollback);
  \draw[refused] (gate) -- node[note, left=0.5mm] {refuse} (rollback);
  \draw[refused] (shacl) |- node[note, pos=0.75, above] {refuse} (rollback);
  \draw[flow] (commit) -- (verdict);
  \draw[flow] (rollback) -- node[note, below] {either way} (verdict);
\end{tikzpicture}}
\caption{The gated write path. Every gate decision --- acceptance or
refusal --- flushes a signed, attributed verdict \emph{after} the
savepoint resolves, so the verdict survives the rollback it records;
the attempted delta deliberately does not.}
\label{fig:writepath}
\end{figure}
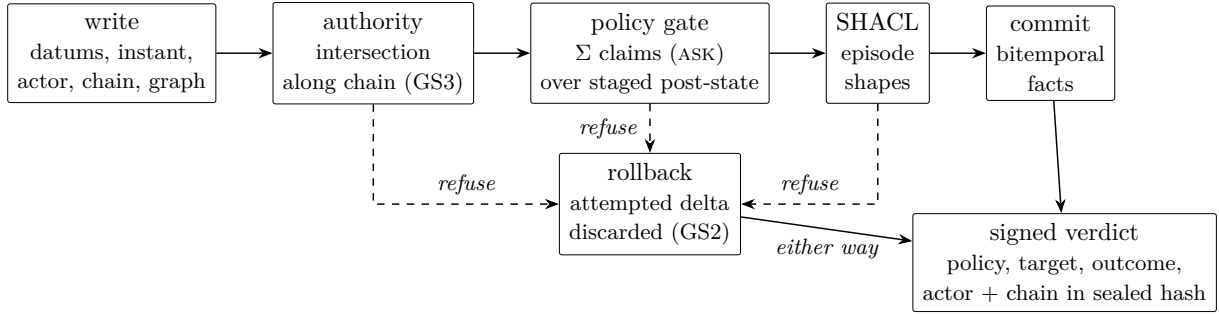

\sys compiles SARC's constraint machinery into the write path, with
\Sig itself stored as facts. Figure~\ref{fig:writepath} shows the
gated write path and the one asymmetry it is built around: a denial
rolls back the attempted delta but not the decision's record.

\paragraph{Policies are facts; the gate is the write path.}
A policy is an entity in the graph: target type, a claim (a SPARQL
\textsc{ask}), a boundary, an effect (\emph{deny},
\emph{require-approval}, advisory effects), and, for escalations, a
declared reversibility window; the full SARC constraint object ---
class and verification point --- is SHACL-validated at \emph{definition}
time, including the class-to-placement discipline plain SHACL cannot
state. At write time, staged facts are evaluated inside the open
savepoint: claims run against the pending post-state, indexed by
target type so a write touching no governed type runs zero claims
(GS1's zero-cost abstention). One consequence surfaced by the
benchmark and worth stating as a rule: a claim's dataset scope is part
of the claim --- a policy governing facts that live in named graphs must
say \textsc{graph} in its pattern, or it judges an empty default-graph
view (\S\ref{sec:census}).

\paragraph{Verdicts survive the rollback they record.}
Every gate decision stages a verdict --- policy, target, outcome, the
write's attribution (actor and principal chain), evidence hash,
signature --- and flushes it in its own transaction \emph{after} the
write's savepoint resolves, so a denial's verdict outlives the
denial's rollback, and keeps its actor even though the attempted
delta does not survive. The attribution sits inside the evidence hash
the signature seals, so ``who'' is not swappable under a valid seal
(\S\ref{sec:eval-demm}). Verdicts are ed25519-signed attestations
against a human-authored verifier registration; a store with no
signing identity records no verdict rather than an unsigned one, and
a re-entry guard prevents a policy targeting verdicts from denying
the recording of its own denial.

\paragraph{Escalation without a waiting engine.}
A refusal under \emph{require-approval} mints a decision request ---
policy, target, evidence hash, expiry from the declared window --- as
facts that survive the refusal's rollback. A human answers with a
decision bound to the same evidence hash; the \emph{next} attempt
succeeds. The hold is the agent retrying, not the engine waiting;
rejection outranks approval; an expired or zero window refuses rather
than inventing a bound.

\paragraph{Authority intersects.}
Authority grants attach principals to graphs. Along a delegation chain
the effective authority is the intersection of every link's; the
wildcard is the identity; an empty intersection refuses with a message
naming the chain, the graph, and what the chain actually holds.
Relabelling a graph requires authority over the meta-graph, not the
graph being labelled.

\paragraph{The audit reads \Sig from the graph.}
The checker decides $T \entails \Sig$ over four passes --- coverage,
class-placement, outcome consistency, attribution --- reading \Sig from
the store rather than a snapshot beside it, deterministically and
never via a model. Violation and incompleteness are kept distinct
end-to-end: the dispatch inventory reports an ungoverned executable
tool class with no declared reason as a violation and the same class
\emph{with} a reason as an acknowledged bypass; an unattributed trace
record is counted incomplete, not placed at the attribution root.
Coverage checking is half-decidable and the audit says which passes
are total.

\section{Implementation}
\label{sec:impl}

\sys is a Rust crate (\raise.17ex\hbox{$\scriptstyle\sim$}67\,kLOC,
1{,}000+ tests) whose store is a single SQLite file --- embeddable with
``SQLite energy,'' servable over a CLI, REST, and MCP tool handlers.
The SPARQL~1.1 evaluator is custom, built over the fact log directly
because off-the-shelf engines assume a mutable triple store with no
temporal columns; it supports the standard operator set plus temporal
parameters (valid-at, as-of-transaction), RDFS widening, and
graph-scoped label reads. Counterfactuals reuse SQLite savepoints:
speculative writes stage, query, and roll back through the same path
the gate uses. Shapes and ontologies live in a bitemporal registry ---
loading a named set closes the prior version rather than overwriting
it, removal is a close rather than a delete, and loads emit events on
the store's append-only spine, so ``the shapes in force at $T$'' is a
lookup (the piece GS6's rules half requires).

\paragraph{Engineering characterization, not comparison.}
We report the store's measured costs as context and make no
comparative query-performance claims; the comparison with conventional
stores in this paper is about defaults and guarantees, not throughput.
On this substrate, storage grows linearly
(\raise.17ex\hbox{$\scriptstyle\sim$}8.3\,KB per ingested episode,
measured from 1k to 20k episodes); write-time SHACL cost is flat in
delta size, dominated by a fixed per-write parse; the evaluator's
basic graph pattern join is a nested loop whose cost grows
quadratically with store size, with an interned-term cache buying a
measured $3.9$--$4.9\times$ constant factor and an in-memory read
model prototype converting a 133\,s two-hop join into 0.15\,ms at
$\sim$385 bytes of resident memory per fact. In the \census runs
(\S\ref{sec:eval}), the gated arm's compliant governed writes carry
roughly $2\times$ the per-write latency of its ungoverned writes
(median 2.7\,ms vs.\ 1.3\,ms in the release-mode seed-42 run),
reflecting claim evaluation plus verdict signing and recording;
ungoverned writes in the gated arm are in turn costlier than the
control arm's (1.3\,ms vs.\ 0.7\,ms) because authority intersection
runs on every graph-scoped write by design --- GS3 is not
abstention-eligible, and RQ1's zero-overhead property is scoped to the
policy gate alone. Single-run latencies are illustrative; the
benchmark's determinism note governs aggregation.

\section{The \census Benchmark}
\label{sec:census}

\census is one scripted, seeded, multi-writer lifecycle over a
governed store --- named for the quipu's original job. A single command
runs one arm; the \emph{gated} arm enforces every gate and the
\emph{control} arm is the same script with the gates off. No LLM runs
anywhere in the core loop: the writers are deterministic drivers, so
the run is its own oracle --- the injector knows every defect it
planted, every metric is a count or a latency, and no judge or rubric
appears anywhere.

\paragraph{Timeline.}
Six phases over three district graphs, three recorder identities with
distinct authority grants, and a scripted human decision role.
\emph{Founding} registers partitions, writers, authority, shapes, and
\Sig (four policies: a label requirement, a single-placement
constraint, a closed-world vocabulary policy requiring every predicate
on a record type to be declared, and an escalation).
\emph{Recording} plants six labeled defects --- an episode missing its
required provenance property, a write into a district outside the
writer's authority, a delegated write exceeding the delegator's grant,
a policy-violating write, a write valid against the \emph{pre}-state
but invalid only in combination (the post-state discriminator), and a
fact using a fabricated predicate --- interleaved with 100 clean writes
(half touching governed types) whose latencies are RQ1's
distributions. \emph{Correction} runs the escalation round-trips
(approve and reject), a supersession, and a trust-plane promotion.
\emph{Composition} runs seven lattice probes: an undeclared member
degrading coverage and failing the floor, a cross-chain trust pair
refusing comparison, an expired label reading as absent, an obligation
joining across the set, clean compositions passing with zero false
refusals, an overlay rebind refused bind-once, and a pack imported
from a second store with a stable content hash. \emph{Amendment}
supersedes one policy's claim mid-run --- \Sig is ordinary bitemporal
facts --- and reloads the record shape through the versioned registry.
\emph{Audit} replays every recorded decision as of its instant, runs
the dispatch inventory over planted tool classes, audits a
synthesized five-record trace with one deliberately unattributed
record, and exports \Sig and the decisions for the external checker.

\paragraph{Reproducibility.}
The only entropy input is the seed (a self-contained SplitMix64);
timestamps are logical minutes from a fixed epoch, so no wall clock
reaches the manifest. Manifests are byte-identical across repeat runs
per (seed, arm) --- measured, three consecutive runs, one hash --- and the
external-checker export is byte-stable. Writing the determinism note
found one real divergence (an absolute output path embedded in an
observed string); it was fixed and the finding kept in place. Scorers
read the manifest --- the injector's declared ground truth --- never the
phase scripts.

\paragraph{In the wild.}
\census is synthetic by construction; to bound the external-validity
gap we recorded five genuine Pre-Action Gate decisions from a real
governed writer (the pre-edit guard of Yupana\footnote{\texttt{github.com/scbrown/yupana} --- the stack's
code-analysis engine and \sys's governed structural writer; formerly
\texttt{hank}.} in enforce mode: two denies
against a blast-radius scope, three allows) and replayed the spool,
unmodified, through the same audit (\S\ref{sec:eval-wild}).

\section{Evaluation}
\label{sec:eval}

All numbers are from the seed-42 \census run (release mode, native
x86-64) unless stated; the manifest, metrics, and set-hashes are in the
artifact, and repeat runs produce byte-identical manifests. Each
research question scores one inversion from Table~\ref{tab:defaults};
Table~\ref{tab:scorecard} is the run at a glance.

\begin{table}[t]
\centering
\footnotesize
\begin{tabular}{@{}lll@{}}
\toprule
 & Question & Result \\
\midrule
RQ1 & enforcement cost scales with governed writes?
    & 0 claims off-target; 1.3 / 2.7\,ms medians (\S\ref{sec:eval}) \\
RQ2 & gated store ends cleaner?
    & \textbf{0 of 6} planted defects vs \textbf{6 of 6} ungated \\
RQ3 & audit decidable; external checker agrees?
    & planted counts exact; SARC agrees verdict-for-verdict \\
RQ4 & composition never widens?
    & \textbf{7 of 7} probes upheld; zero false refusals \\
RQ5 & decisions replay as of their instant?
    & \textbf{50 of 50} re-derive; 6 of 6 denials verify rules-in-force \\
\midrule
ext. & evidence sufficient, reader never overclaims?
    & PSA \textbf{1.0}, overclaim \textbf{0.0} (\S\ref{sec:eval-demm},
      Table~\ref{tab:demm}) \\
\bottomrule
\end{tabular}
\caption{The evaluation at a glance: five research questions scored by
the seeded \census run against planted ground truth, plus the external
decision-evidence sufficiency benchmark over the run's exported
records.}
\label{tab:scorecard}
\end{table}

\subsection{RQ1 --- does enforcement cost scale only with governed writes?}
Partly, and the boundary is itself a result. Within the gated arm,
clean writes touching no governed type run zero policy claims (the
target-type pre-filter) and median 1.3\,ms per write versus 2.7\,ms
for compliant governed writes, which pay claim evaluation plus verdict
signing and recording. But the gated arm's ungoverned writes are not
free relative to the control arm's (median 0.7\,ms): authority
intersection runs on every graph-scoped write \emph{by design} --- GS3
guards where facts may land at all and is deliberately not
abstention-eligible. Zero-cost abstention is a property of the policy
gate, not of governance as a whole, and we report it at that scope.

\subsection{RQ2 --- does the gated store end cleaner than the ungated one?}
Completely, on the planted set: the gated arm's final graph contains
\textbf{0 of 6} planted defects; the control arm --- the same script,
gates off --- contains \textbf{6 of 6}. Each refusal names its ground:
the SHACL episode gate names the missing provenance property; the
authority refusals name the chain, the graph, and what the chain
actually holds; the policy denial cites the policy; the post-state
probe (CEN-P2) is refused by the combined-state claim a pre-state gate
would pass; and the fabricated predicate is refused by the
closed-world vocabulary policy, which is not new machinery but an
ordinary \textsc{ask} claim in \Sig requiring every predicate on the
record type to be declared.

\subsection{RQ3 --- is the audit decidable in-store, and does an external
checker agree?}
\label{sec:eval-audit}
In-store, the four passes produce exactly the planted distinctions:
the dispatch inventory reports \textbf{1 violation} (an executable
tool class, ungoverned, no reason) and \textbf{1 incompleteness} (the
same shape \emph{with} a declared reason --- an acknowledged bypass, not
an unknown hole); the trace audit over a five-record window reports
\textbf{0 violations and 1 incompleteness} --- the deliberately
unattributed record, counted incomplete rather than misreported or
placed at the attribution root.

Against the SARC reference checker
(\texttt{besanson/sarc-governance}), we export \Sig in its spec format
and the run's 56 decisions in its flat-trace format, twice. The
\emph{faithful} export --- only the evaluations \sys actually ran ---
fails the reference checker with 168 discrepancies, \textbf{all of
type coverage and none of verdict, placement, or response}: the
reference invariant expects every constraint evaluated for every
action, and \sys's target-type pre-filter is invisible to it. The
\emph{padded} export, which adds explicit not-fired records for the
non-applicable constraints, passes clean. The two checkers agree
verdict-for-verdict and disagree only on coverage \emph{semantics} ---
whether abstention must be materialized to be auditable --- which we
report as the finding rather than resolving by fiat.

\subsection{RQ4 --- does composition never widen?}
All \textbf{7 of 7} composition probes uphold the lattice contract:
the undeclared member degrades coverage to \textsf{Partial} and the
enforcement floor refuses the composed read; the cross-chain trust
pair is refused with both chains named; the expired label reads as
absent and degrades coverage; the \texttt{no-export} obligation joins
across the set; all four clean compositions pass the floor ---
\textbf{zero false refusals}; the overlay rebind is refused bind-once;
and the imported pack re-interns with its content hash stable across
id spaces.

\subsection{RQ5 --- do decisions replay as of their instant?}
\label{sec:eval-replay}
The amendment supersedes the label policy's claim mid-run, so every
phase-2 decision predates a rule change. Replaying all 56 recorded
decisions: \textbf{50 of 50} satisfied verdicts re-derive faithfully
under the claim and data as of their instant --- and \textbf{all 50}
evaluate \emph{unsatisfied} under the amended claim, the number that
separates ``the runtime got it wrong'' from ``the spec moved,'' and
exactly what a latest-only rule set would misreport. The
\textbf{6} denials replay differently, and honestly so: their staged
deltas were rolled back --- GS2 keeps the verdict, deliberately not the
attempt --- so replay verifies the rules in force at the instant
(6 of 6) rather than re-deriving the outcome. Full denial
re-derivation would require traces to carry the attempted delta,
which the governed writer's traces do and the store deliberately does
not; the asymmetry is a property of refusal-by-rollback, not a gap in
bitemporality.

\subsection{The agent arm: strictness borne by the writer}
\label{sec:eval-agent}
An LLM agent is given the recording task cold --- five actions, no
sight of \Sig, the gate's verbatim refusals as the only feedback
channel --- and gets one revision. The original run (single model,
$n{=}1$, transcripts committed) went 2 accepted / 3 refused, then
\textbf{5 of 5}. We repeated the protocol across four Claude models,
three trials each (twelve trials; prompt and all transcripts
committed under \texttt{benchmark/census/agent/} in the archived
artifact~\cite{brown2026quipu}), with the scenario
arranged so one refusal is \emph{not} fixable by editing the record:
household h3 belongs to a district the writer's authority does not
reach.

\begin{table}[t]
\centering
\small
\begin{tabular}{@{}llll@{}}
\toprule
Model ($\times$3) & Attempt 1 & Attempt 2 & The unfixable refusal's resolution \\
\midrule
haiku-4.5 & 4/5, 2/5, 2/5 & all accepted & dropped $\times$2; refiled north $\times$1 (false, landed) \\
sonnet-5  & 4/5 $\times$3 & 4/5 $\times$3 & parked in root $\times$3 --- re-refused (\textsc{graph} scope) \\
opus-5    & 4/5 $\times$3 & 4/5 $\times$3 & resubmitted unchanged, caveats written into the record \\
fable-5   & 4/5 $\times$3 & 5/5, 5/5, 4/5 & routed via the authorized chain $\times$1; root $\times$2 \\
\bottomrule
\end{tabular}
\caption{The agent arm across models, three trials each. Every
label/vocabulary refusal was fixed in one revision; the authority
refusal, unfixable by editing, split the models into routing,
principled abstention, evasion-caught, and evasion-landed.}
\label{tab:agentarm}
\end{table}

Three regularities (Table~\ref{tab:agentarm}). \emph{Everything \Sig
can name converges}: all label and vocabulary refusals were fixed in
one revision in 12 of 12 trials, no trial invented a predicate, and
the cheapest model gained the most --- haiku went 2/5 to full
acceptance on its weakest attempts, which is where a gate that
explains its refusals pays best. \emph{The authority refusal sorts
writers by disposition, not capability}: one fable trial made the
correct move (dispatch h3 through the chain that holds the south);
all three opus trials declined every workaround as falsification and
wrote their caveats into the record itself --- refusing to trade truth
for acceptance; five trials tried to park the record in a graph the
writer does hold, and the gate caught every one, because the
tally-label claim is \textsc{graph}-scoped and a tally in the default
graph fails it. \emph{The gate is exactly as good as \Sig}: the one
false record that landed (h3 refiled to the northern district) passed
because no policy states which district a household belongs to ---
the residual risk under refusal-driven convergence concentrates
precisely in what \Sig leaves unsaid, which is an argument for
authoring coverage, not against the gate. SARC-DQ~\cite{besanson2026sarcdq}
reports the ungated complement of the same design: agents handed
defective evidence convert it into costly actions at a rate flat
across four model tiers spanning ${\sim}15\times$ in price ---
capability does not buy skepticism --- so the convergence observed here
is the gate's doing, not the models'. Boundaries: one task, one
model family, three trials per model, and the scripted scenario as
the only quality oracle.\footnote{A richer oracle is in progress:
camayoc (\texttt{github.com/scbrown/camayoc}) maintains the stack's
competency-question suites, and scoring agent-written facts against
those suites would replace the scripted scenario here.}

\subsection{In the wild}
\label{sec:eval-wild}
Replaying the recorded Yupana trace through the audit, with the shipped
policy catalog as \Sig, yields $T \not\entails \Sig$: \textbf{2
violations, 6 incompleteness findings} over 5 records. The violations
are real and actionable --- the guard enforced a locally-configured
blast-radius rule that is not an authored policy in \Sig, and the
finding's remediation is the thesis in one line: \emph{author it in
the store so it can be audited, or stop enforcing it}. The
incompleteness findings are the honest remainder: two catalog
constraints the window never exercised, undeclared placement on the
config-file rule, and partial attribution the runtime never recorded.
The synthetic \census confirms planted defects are caught; the wild
trace shows the same machinery surfacing an unstaged enforcement gap.

\subsection{Evidence sufficiency against an external benchmark}
\label{sec:eval-demm}
DEMM-Bench~\cite{solozobov2026demmbench} asks the converse of RQ3: not
whether decisions were \emph{correct}, but whether the records a
runtime emits \emph{suffice to reconstruct} eight decision-level
properties (actor identity, principal authority, action boundary,
policy basis, decision basis, data/resource touch, lifecycle context,
verification strength) under eight controlled degradation conditions.
Its lead diagnostic is \emph{Overclaim Rate} --- declaring a decision's
evidence ``sufficient'' while a required property is not
reconstructable --- which operationalises the \emph{container fallacy}:
inferring audit sufficiency from the presence of a trace, ledger, or
schema. Following the benchmark's invitation to add regimes and
scorers on its published contracts, we run \sys as a self-added ninth
regime (\texttt{benchmark/demm/}): the \census run exports its 56
recorded decisions as three-plane native records --- writer-side guard
trace, signed verdict fact queried back from the store, bitemporal
policy snapshot --- our transforms apply the benchmark's eight
degradation semantics as content-level deletions, and a content-only
reader reconstructs the eight properties, scored against the
benchmark's own construction oracle over 64 cases. The degradations
and the adapter are ours, so the result is a claim about \sys's
record \emph{format}, not a leaderboard comparison against the
benchmark's deliberately redacted-input reference scorer.

\begin{figure}[t]
\centering
\resizebox{\textwidth}{!}{%
\begin{tikzpicture}
  % Stacked relative to the centre node rather than at fixed y-coordinates:
  % three lines of text are taller than the 1.35cm spacing allowed, so the
  % boxes overlapped. `below=of` guarantees the gap whatever the content.
  \node[stage] (ledger) at (0, 0) {signed verdict\\\footnotesize policy, outcome,\\\footnotesize attribution, hash, sig};
  \node[stage, above=3mm of ledger] (guard) {guard trace\\\footnotesize writer, chain, tool,\\\footnotesize target, graph, instant};
  \node[stage, below=3mm of ledger] (policy) {policy snapshot\\\footnotesize claim as-of vs now,\\\footnotesize authority grants};
  \node[draw, dashed, rounded corners=1pt, fit=(guard)(ledger)(policy),
        inner sep=3pt, label={[note]above:{56 \census decisions $\times$ 3 planes}}] (native) {};
  \node[stage] (degrade) at (4.1, 0) {8 degradations\\\footnotesize delete or contradict\\\footnotesize evidence content};
  \node[stage] (presence) at (8.0, 1.2) {presence baselines\\\footnotesize is a container there?\\\footnotesize\itshape scores 0.875 overclaim};
  \node[stage] (reader) at (8.0, -1.2) {property-level reader\\\footnotesize what does the content\\\footnotesize still establish?\\\footnotesize\itshape scores PSA 1.0, 0 overclaim};
  \node[stage] (oracle) at (12.3, 0) {construction oracle\\\footnotesize benchmark's own\\\footnotesize ground truth};
  \draw[flow] (native) -- (degrade);
  \draw[flow] (degrade.east) -- (presence.west);
  \draw[flow] (degrade.east) -- (reader.west);
  \draw[flow] (presence.east) -- (oracle.west);
  \draw[flow] (reader.east) -- (oracle.west);
\end{tikzpicture}}
\caption{The DEMM run (\texttt{benchmark/demm/}): \census exports each
decision as three evidence planes; deterministic degradations damage
the content; two kinds of reader are scored against the benchmark's
construction oracle over 64 cases (8 conditions $\times$ 8 question
families).}
\label{fig:demm}
\end{figure}
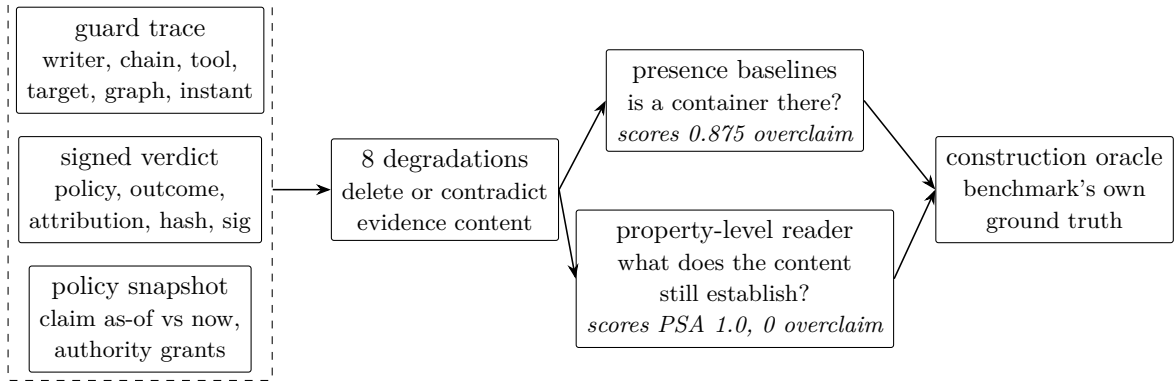

\begin{table}[t]
\centering
\small
\begin{tabular}{@{}lrrrr@{}}
\toprule
Reader of \sys's evidence & Sufficient & Overclaim & Underclaim & PSA \\
\midrule
property-level reader           & 8/64  & \textbf{0.000} & 0.000 & \textbf{1.000} \\
\sys-internal validity check    & 8/64  & 0.000 & 0.000 & --- \\
container checklist (3 planes)  & 56/64 & 0.750 & 0.000 & --- \\
trace- / ledger- / schema-present & 64/64 & 0.875 & 0.000 & --- \\
\midrule
\emph{always-sufficient anchor} & 64/64 & 0.875 & 0.000 & --- \\
\emph{always-insufficient anchor} & 0/64 & 0.000 & 1.000 & --- \\
\bottomrule
\end{tabular}
\caption{DEMM-Bench over \sys's exported evidence (64 cases; 512
property judgements). Reference points on the benchmark's own corpus:
presence baselines overclaim 0.50--0.75, its stricter validators
0.00, and its redacted-input candidate scorer reaches PSA 0.5625 at
zero overclaim. Comparisons across corpora are indicative only; the
within-corpus contrast between rows is the finding.}
\label{tab:demm}
\end{table}

Three findings, all deterministic
(Figure~\ref{fig:demm}, Table~\ref{tab:demm}). First, the container
fallacy reproduces on \sys's evidence \emph{at its ceiling}: trace-,
ledger-, and schema-presence baselines overclaim on 87.5\% of cases
(the benchmark's own corpus: 50--75\%), precisely because \sys always
emits all three planes --- content-level degradation leaves every
container present, so presence carries no information at all. Second,
mechanical validity checking is enough to stop overclaim but not to
localise it: a \sys-internal validator (field completeness,
evidence-hash recomputation, executor--chain consistency, grant
scope) reaches zero overclaim by refusing every degraded record
outright, matching the benchmark's stricter baselines. Third, the
property-level reading reconstructs \textbf{512 of 512} property
cells (mean PSA 1.0, zero overclaim, zero underclaim) --- including
the two slices the benchmark reports as its candidate's hardest
(conflicting identity and action boundary, PSA 0.25 there): both are
decidable from \sys's records because the guard trace names its
tool, target, graph, and principal chain, and the signed verdict
names its policy, whose claim the store serves as of the decision
instant. What a degradation deletes is detectable \emph{as} absent,
so sufficiency never has to be guessed from container presence.

\paragraph{Impartiality by construction, not by administration.}
The run is self-administered, and its impartiality rests on the same
discipline as the \census itself, not on who pressed the button.
Ground truth is the benchmark's own published construction oracle,
unmodified; the degradation transforms and the reader are published
in full alongside the store, and the reader consumes evidence content
only --- case identifiers are opaque, no degradation name reaches any
scorer-facing field, and the benchmark's label-leakage rules are
honoured. Every stage is deterministic: reruns from the census binary
up reproduce byte-identical artifacts, so the numbers in
Table~\ref{tab:demm} are a property of the published pipeline, not of
a run we witnessed. Anyone can re-execute the whole chain --- seeded
census, export, degradation, scoring --- from the repository in
minutes, substitute their own reader for ours on the same manifests,
or tighten the transforms and re-score; the claim survives exactly as
far as those reruns agree.

The run also changed the store. Its first pass surfaced two evidence
gaps the benchmark's vocabulary made precise: the verdict signature
sealed the outcome but not the writer binding, so a conflicting
identity was catchable only by a consistency convention; and a
denial's actor was not persisted at all --- GS2 rolls the attempt back,
so ``who was refused?''\ was answerable only from writer-side traces
the store does not own, the same division of labour the wild-trace
audit (\S\ref{sec:eval-wild}) reported as partial attribution. Both
closed in one change (Q-VERDICT-ATTRIB): the verdict fact now carries
the write's attribution --- actor and principal chain --- inside the
evidence hash its signature seals, so a swapped writer fails hash
recomputation rather than needing a rule to notice, and a refusal
keeps its actor while the attempted delta still does not survive.
The numbers above are from the improved store; the benchmark
re-verifies them end-to-end, since the validator's recomputation of
the extended hash is what certifies the eight intact cases.

\section{Related Work}
\label{sec:related}

\paragraph{Bitemporal stores.}
Datomic established the immutable EAVT log with transaction time;
XTDB adds valid time as a first-class axis. \sys's departure is not
the axes but their reach: labels, verdicts, decisions, authority
grants, and --- via the versioned registry --- the validation rules
themselves are time-indexed, which is what makes governance replay
(\S\ref{sec:eval-replay}) a query rather than an archaeology project.

\paragraph{Named graphs, provenance, and information flow.}
Carroll et al.~\cite{carroll2005named} attached provenance and trust
to named graphs; annotated-RDF and semiring-provenance lines
formalized per-triple annotations; Denning's
lattice~\cite{denning1976lattice} is the ancestor of composing
security-relevant labels by meet and join. \sys's contribution on
this line is operational: coverage as an explicit third outcome
(neither fail-open $\top$ nor floor-dragging $\bot$), refusal of
cross-chain comparison by name, expiry-as-absence, and a
machine-checked composition homomorphism --- enforced at query time by
floors rather than described.

\paragraph{Validation-centric stores.}
SHACL provides shape validation, and stores ship it as a callable
service; \sys makes it (and policy claims) a write-path gate with
episode scope, and pairs open-world shape validation with closed-world
vocabulary policies expressed as ordinary \textsc{ask} claims ---
catching the fabricated-term case open-world validation silently
accepts, a failure mode measured at scale by
Rovai~\cite{rovai2026openontologies}.

\paragraph{Governance for agentic systems.}
SARC~\cite{besanson2026sarc} is the nearest ancestor and the frame we
compile: constraints as specification objects, four enforcement
points, decidable audit. \sys relocates the machinery into the store ---
\Sig as facts, the gate as the write path, verdicts as signed
bitemporal facts, the audit reading \Sig from the graph it governs ---
and \S\ref{sec:eval-audit} measures agreement with SARC's own
reference checker. SARC's successors specialise the same frame per
domain: Green SARC~\cite{besanson2026greensarc} compiles cost and
carbon budgets into the loop (a soft budget penalty breaches on
91.5\% of seeds, the architectural gate on none --- the
accept-then-clean lesson restated in FinOps), and
SARC-DQ~\cite{besanson2026sarcdq} gates evidence quality at the point
of action, the loop-side counterpart of our write gate; \sys is the
specialisation to the store. Verification runtimes such as
Tardygrada~\cite{rovai2026tardygrada} converge independently on
three-valued verdicts and weakest-link aggregation, evidence that the
violation/incompleteness distinction and least-confident-leaf rules
are being rediscovered wherever agent output must be gated. The same
migration is under way in practitioner harnesses:
ForCoding~\cite{forcoding2026} rebuilt its orchestrator by
consolidating sixty-odd prompt-embedded rules --- ``polite requests''
to the model, by its own changelog --- into a thirty-one-rule
deterministic policy engine with hash-chained audit records: the governance-outside default being
abandoned in the field, at the harness layer, for the same reasons we
abandon it in the store.

\paragraph{LLM-driven knowledge-graph construction.}
Tool-augmented ontology engineering~\cite{rovai2026openontologies,
zhang2024accelerating} is the workload that motivates strictness:
agents produce plausible, well-formed, sometimes fabricated structure
at rates no curation pass matches. Where that line builds better
construction tools, we build the store those tools should be made to
convince.

\section{Conclusion}
\label{sec:conclusion}

The conventional knowledge graph was designed for a writer who no
longer writes it alone. We presented \sys, a store that inverts the
four defaults that assumption baked in --- refuse at the gate,
bitemporal everything, partitioned trust with non-widening
composition, governance inside --- and \census, a deterministic
lifecycle that measures each inversion against planted ground truth
and against a real writer's recorded trace. The measured story is
consistent: strictness is affordable when its cost falls on agents who
can read a refusal and retry; refusals are worth recording as signed
facts precisely because they are the events audit needs; composition
can be made safe without being made silent; and a store whose rules
are as historical as its data can replay not just what it knew but
what it required. An external sufficiency benchmark run over the
store's own exported evidence closes the loop from the outside:
every property-level governance question stays answerable exactly
when the evidence warrants it, with zero overclaim --- and the run's
first pass improved the store it measured, sealing attribution into
the signed verdict.

Three boundaries are stated rather than hidden: no comparative
query-performance claims (the evaluator's ceiling is characterized,
not raced); labels are not access control (a floor refuses a query, it
does not hide rows); and denials replay as attestation checks, not
re-derivations, because refusal-by-rollback deliberately keeps the
verdict and discards the attempt. Future work runs in two directions:
the controlled agent-arm experiment (multiple tasks and models, with a
competency-question oracle) and lifting GS1--GS6 from design principles
to a portable contract --- a governed-store specification other
substrates can claim and this benchmark can score, extending SARC's
compilation one layer further down for stores we have not built.

\bibliographystyle{plain}
\bibliography{references}

\begin{thebibliography}{10}

\bibitem{besanson2026greensarc}
Gaston Besanson.
\newblock Green {SARC}: Predictive cost and carbon governance for agentic {AI}
  systems.
\newblock {\em arXiv preprint arXiv:2606.15954}, 2026.

\bibitem{besanson2026sarc}
Gaston Besanson.
\newblock {SARC}: A governance-by-architecture framework for agentic {AI}
  systems: Compiling regulatory obligations into runtime constraints.
\newblock {\em arXiv preprint arXiv:2605.07728}, 2026.
\newblock Reference artifacts: \texttt{github.com/besanson/sarc-governance}.

\bibitem{besanson2026sarcdq}
Gaston Besanson.
\newblock {SARC-DQ}: Runtime data-quality gating for agentic {AI}: Silent
  evidence defects, the incompetence shield, and downstream-only remediation.
\newblock {\em arXiv preprint arXiv:2607.26313}, 2026.

\bibitem{brown2026quipu}
Steve Brown.
\newblock {Quipu}: an {AI}-native knowledge graph with strict ontology
  enforcement, 2026.
\newblock Archived source and \texttt{benchmark/census/} artifacts for every
  number reported here. Concept DOI \texttt{10.5281/zenodo.21878428} resolves
  to the newest release.

\bibitem{carroll2005named}
Jeremy~J. Carroll, Christian Bizer, Pat Hayes, and Patrick Stickler.
\newblock Named graphs, provenance and trust.
\newblock In {\em Proceedings of the 14th International Conference on World
  Wide Web (WWW)}, pages 613--622, 2005.

\bibitem{denning1976lattice}
Dorothy~E. Denning.
\newblock A lattice model of secure information flow.
\newblock {\em Communications of the ACM}, 19(5):236--243, 1976.

\bibitem{forcoding2026}
{ForCoding Contributors}.
\newblock Forcoding: a policy-enforced agent orchestrator plugin for opencode.
\newblock \url{https://github.com/devrockin/forcoding}, 2026.

\bibitem{rovai2026openontologies}
Fabio Rovai.
\newblock Open ontologies: Tool-augmented ontology engineering with stable
  matching alignment.
\newblock {\em arXiv preprint arXiv:2605.09184}, 2026.

\bibitem{rovai2026tardygrada}
Fabio Rovai.
\newblock Tardygrada: a verification runtime for agent outputs.
\newblock \url{https://github.com/fabio-rovai/tardygrada}, 2026.

\bibitem{solozobov2026demmbench}
Oleg Solozobov.
\newblock {DEMM-Bench}: A cross-regime benchmark for agent-runtime
  governance-evidence sufficiency.
\newblock {\em arXiv preprint arXiv:2606.20634}, 2026.
\newblock Reference artifacts:
  \texttt{github.com/agent-runtime-evidence/decision-evidence-benchmark}.

\bibitem{zhang2024accelerating}
Bohui Zhang et~al.
\newblock Accelerating knowledge graph and ontology engineering with large
  language models.
\newblock {\em arXiv preprint arXiv:2411.09601}, 2024.

\end{thebibliography}

\end{document}